\documentclass[lettersize,journal]{IEEEtran}
\usepackage{amsmath,amsfonts,amssymb}
\usepackage[ruled]{algorithm2e}
\usepackage{array}
\usepackage{bbm}
\usepackage[caption=false,font=normalsize,labelfont=sf,textfont=sf]{subfig}
\usepackage{textcomp}
\usepackage{stfloats}
\usepackage{url}
\usepackage{verbatim}
\usepackage{graphicx}
\usepackage{booktabs}
\usepackage{multirow} 
\usepackage{cite}
\usepackage{hyperref} 

\usepackage{siunitx}
\usepackage{color}
\usepackage{tabularx}

\makeatletter

\newcommand{\Rmnum}[1]{\expandafter\@slowromancap\romannumeral #1@}
\makeatother

\begin{document}
%

\title{
Human-Like Implicit Intention Expression for Autonomous Driving Motion Planning: A Method Based on Learning Human Intention Priors
}

\author{Jiaqi Liu,~\IEEEmembership{Student Member,~IEEE,} Xiao Qi, Ying Ni, Jian Sun, and Peng Hang,~\IEEEmembership{Member,~IEEE}
\thanks{This work was supported in part by the National Key R\&D Program of China (2022YFB2502901), the National Natural Science Foundation of China (52125208, 52232015, 52272313), the Young Elite Scientists Sponsorship Program by CAST (2022QNRC001) and the Fundamental Research Funds for the Central Universities.}
\thanks{Jiaqi Liu,  Xiao Qi, Ying Ni, Jian Sun and Peng Hang  are with the Department of
Traffic Engineering and Key Laboratory of Road and Traffic Engineering,
Ministry of Education, Tongji University, Shanghai 201804, China. (e-mail: \{liujiaqi13, qixaio, ying\_ni, sunjian, hangpeng \}@tongji.edu.cn)}

\thanks{Corresponding author: Peng Hang}
}

\markboth{}%
{Shell \MakeLowercase{\textit{et al.}}: A Sample Article Using IEEEtran.cls for IEEE Journals}

\maketitle 

\begin{abstract}
One of the key factors determining whether autonomous vehicles (AVs) can be seamlessly integrated into existing traffic systems is their ability to interact smoothly and efficiently with human drivers and communicate their intentions. While many studies have focused on enhancing AVs' human-like interaction and communication capabilities at the behavioral decision-making level, a significant gap remains between the actual motion trajectories of AVs and the psychological expectations of human drivers. This discrepancy can seriously affect the safety and efficiency of AV-HV (Autonomous Vehicle-Human Vehicle) interactions. To address these challenges, we propose a motion planning method for AVs that incorporates implicit intention expression. First, we construct a trajectory space constraint based on human implicit intention priors, compressing and pruning the trajectory space to generate candidate motion trajectories that consider intention expression. We then apply maximum entropy inverse reinforcement learning to learn and estimate human trajectory preferences, constructing a reward function that represents the cognitive characteristics of drivers. Finally, using a Boltzmann distribution, we establish a probabilistic distribution of candidate trajectories based on the reward obtained, selecting human-like trajectory actions. We validated our approach on a real trajectory dataset and compared it with several baseline methods. The results demonstrate that our method excels in human-likeness, intention expression capability, and computational efficiency.
\end{abstract}


\section{Introduction}
The emergence of autonomous vehicles (AVs) marks a revolutionary era in transportation, promising widespread transformation across the entire system. In the dynamic landscape of mixed traffic scenarios, ensuring the harmonious coexistence of autonomous vehicles and human-driven vehicles (HVs) is crucial~\cite{toghi2021cooperative,wang2022social}.
A key factor in integrating AVs into existing traffic systems is their ability to interact smoothly and intelligibly with human drivers. Unlike human drivers, who can engage in flexible and diverse social interactions, current AVs still lack effective communication and interaction capabilities\cite{toghi2022social,liu2024enhancing}. In recent years, enhancing the social interaction abilities of AVs has become a focal point for many researchers\cite{schwarting2019social,crosato2022interaction}. However, most studies have concentrated on the behavioral decision-making level. While AVs can exhibit good interactive abilities in terms of decision-making intentions, there remains a significant gap between the actual motion trajectories of AVs and the expectations of human drivers.

In real-world traffic scenarios, motion trajectories serve as a critical means for human drivers to interact and communicate. For instance, drivers can subtly express their intentions during interactions through slight trajectory shifts and speed adjustments\cite{de2013road,lee2021road}. However, existing AV motion planning algorithms often fail to account for these factors. For example, traditional discrete optimization-based algorithms can satisfy requirements such as ease of construction, smooth trajectories, and continuous differentiability. These algorithms are widely used in AV motion planning, enabling simple trajectory actions like collision avoidance and lane changes in high-speed following and lane-changing scenarios. Nevertheless, they struggle to accurately mimic the real driving trajectories of human drivers, leading to significant discrepancies between planned and actual human trajectories. When humans interact with AVs controlled by such planning algorithms, these discrepancies can result in mismatched psychological expectations, thereby affecting the efficiency and safety of interactions \cite{wang2022social,valiente2023prediction}.


In response to these issues, some researchers, such as Huang\cite{huang2021driving} and Song\cite{Song1797}, have introduced learning-based trajectory planning techniques aimed at mimicking human-like behavior by drawing inspiration from and imitating expert human trajectories. While these methods have the potential to utilize real HV trajectories to guide AV trajectory planning, they still leave several critical challenges unresolved. One key challenge lies in the fact that high-level decision intentions are crucial in constraining the trajectory planning space. Existing learning methods typically derive AV decisions from trajectory outcomes, which contradicts the fundamental principle of 'decision before execution'. Moreover, these methods often overlook the intention expression embedded in trajectory movements influenced by high-level decisions. This oversight is particularly evident in complex intersection scenarios where different trajectories are involved.

The intention information embedded within trajectory movements plays a crucial role in interactions, leading to significant differences in trajectory strategies and expected spaces under varying decisions. This complexity and importance are further heightened when AVs navigate intersections, which are among the most demanding interaction scenarios. Such complexity often hinders AVs from executing unprotected left turns with the finesse of a human driver\cite{zhao2023unprotected}.

To address these challenges, we propose an AV motion planning framework that incorporates implicit intention expression. This framework learns prior intentions from human driving data and constructs expected trajectory space constraints based on these learned intentions, generating candidate motion trajectories that account for intention expression. This process enables the compression and pruning of the trajectory space.
We then design a reward function that considers efficiency, comfort, and safety to represent the cognitive characteristics of drivers. Maximum Entropy Inverse Reinforcement Learning (ME-IRL) is employed to learn and evaluate human trajectory preferences, and we establish a probability distribution for trajectory selection influenced by the Boltzmann distribution to simulate the human decision-making process.
We applied and tested our method using the most representative unprotected left-turn task at unsignalized intersections. Using a real-world trajectory dataset, we validated our algorithm in both a simulation environment and a human-in-the-loop driving simulator. Comparisons with several baseline algorithms demonstrate that our method excels in human-likeness, intention expression capability, and computational efficiency.

In summary, our contributions are shown as follows:
\begin{itemize}
    \item We propose an AV motion planning method with implicit intention expression, capable of generating motion trajectories that consider intention expression by learning prior knowledge from human trajectories.
    \item Maximum Entropy Inverse Reinforcement Learning is used to learn and evaluate human driving trajectory preferences, with a reward function designed to account for driver cognitive characteristics.
    \item Our method has been validated in both simulation and human-in-the-loop driving environments, with experimental results showing superior overall performance.
\end{itemize}

The rest of the paper is organized as follows: Section~\ref{sec:2} summarizes the recent related works. 
In section~\ref{sec:3}, the trajectory planning approach we proposed is described. In section \ref{sec:4}, the  experiments are introduced and the results are analyzed. Finally, this paper is concluded in section \ref{sec:5}.

\section{Related Works}
\label{sec:2}
\subsection{Intention Expression Between Vehicles}
Research has demonstrated that intention expression and communication between drivers can significantly enhance traffic efficiency and safety, especially in complex driving scenarios such as unsignalized intersections without clear right-of-way\cite{lee2021road,de2013road}. These forms of intention interaction typically include both explicit and implicit communication methods\cite{de2013road}. Explicit methods involve drivers actively sending or displaying specific signals or information to other vehicles, such as gestures, flashing lights, honking, or eye contact. Implicit methods, on the other hand, involve vehicles subtly conveying their intentions or status to others through their driving behavior, such as speed changes, trajectory shifts, or following distance. Human drivers generally prefer and trust implicit communication methods for completing interactive driving tasks.

Studies have also found that AVs can improve interaction efficiency and safety with human drivers by implicitly expressing intentions through adjustments in approach speed, distance, lateral displacement, and braking position. Rettenmaier et al. \cite{rettenmaier2021communication} investigated the impact of AV speed and lateral displacement on vehicle interaction in bottleneck road sections, and their results showed that when AVs included lateral displacement in their movements, participants experienced better traffic efficiency and higher safety. Similarly, left-turning vehicles at intersections can use implicit communication methods such as early lateral displacement\cite{YangLan2022Human}, delayed lateral displacement\cite{ma2017two2}, or early yaw angle deviation\cite{rettenmaier2021communication} in their trajectories to signal whether they intend to proceed or yield, making their intentions clearer and easier for the other party to understand.
While the above studies have explored the effects of implicit intention expression, few have considered incorporating intention expression into trajectory planning. This oversight means that AVs' planned motion trajectories might confuse human drivers, potentially compromising safety and efficiency.

\subsection{Trajectory Planning in Autonomous Vehicle Systems}
The development of trajectory planning algorithms is fundamental in the hierarchical structure of autonomous driving systems, especially post the establishment of decision intent during motion. These algorithms are essential for generating paths that are not only collision-free but also executable \cite{hang2023brain}.

A prevalent approach in trajectory planning is the use of discrete optimization methods, exemplified by the Frenet trajectory planning method \cite{werling2010optimal}. This method relies on the Frenet coordinate framework, simplifying the trajectory planning into two-dimensional spatial-temporal (S-T) and lateral-temporal (L-T) problems. Building upon this, various researchers \cite{zhang2020optimal,hu2022probabilistic,hu2018dynamic} have sought to refine and optimize the Frenet method. For instance, the team at Apollo \cite{zhang2020optimal} adapted the Frenet approach to devise a quadratic optimization strategy, accommodating the non-holonomic constraints of vehicles. Hu et al. \cite{hu2018dynamic} developed a dynamic path planning method, incorporating the Gaussian convolution algorithm to evaluate collision risks with both static and dynamic obstacles.

Alternative trajectory characterization techniques, such as the Archimedean spiral \cite{ma2017two} and the Bezier curve \cite{zhou2022autonomous}, are also employed, particularly useful in scenarios like intersection turns or high-speed lane changes. These methods, however, demand additional parameters, including control points and parameters for trajectory. Unlike polynomial methods, which inherently integrate linear speed and acceleration outcomes, these techniques require further resolution of motion planning parameters.

Compared to direct motion planning methods that yield discrete kinematic control parameters instantaneously, discrete optimization-based trajectory planning algorithms offer the advantage of generating and planning rational operational trajectories within a controlled generation space. This approach effectively addresses challenges such as the absence of comprehensive planning, trajectory smoothness, and the overlooking of trajectory diversity and the crucial aspect of intent expression.

While discrete optimization-based trajectory planning methods can satisfy basic motion planning needs, significant disparities persist between its planned trajectories and human trajectories\cite{huang2021driving}. Moreover, its inability to express intent impedes effective interaction between Avs and HVs\cite{wang2022social}. In our work, the intention-expressing ability will be added into the trajectory planning framework to enhance the social interaction performance of AVs.

\subsection{Inverse Reinforcement Learning for Motion Planning}

In numerous existing studies \cite{hu2022probabilistic,werling2010optimal}, feature weights in the reward function of trajectory planning are assigned manually or optimized through numerous experiments, resulting in a lack of alignment with driver cognitive characteristics. Inverse Reinforcement Learning (IRL) offers a solution to this issue by reconstructing the reward function through learning from expert example trajectories \cite{arora2021survey}. IRL has been extensively applied to problems such as route planning \cite{ziebart2008maximum,wulfmeier2017large} and the learning of human behavior trajectory features \cite{abbeel2004apprenticeship}. Notable IRL methods include Maximum Entropy IRL \cite{ziebart2008maximum,wulfmeier2017large}, Apprentice IRL \cite{abbeel2004apprenticeship}, and Bayesian IRL \cite{ramachandran2007bayesian}. Among these, ME-IRL holds substantial benefits in studying human expert trajectory behavior, including uncertainty modeling, learning behavior diversity, and robust generalization ability.

Some studies have combined discrete optimization-based trajectory planning methods with ME-IRL to learn human driving behavior and trajectory selection mechanisms. Wu et al. \cite{wu2020efficient} proposed a sampling-based continuous-domain ME-IRL algorithm that improves sampling efficiency through spatiotemporal decoupling and adaptive sampling, effectively learning human driving behavior. Huang et al.\cite{hang2020human} further incorporated vehicle-to-vehicle interactions by describing trajectory distribution using Boltzmann noise theory, assuming an exponential relationship between trajectory probability and reward, and ultimately developed a highway driving trajectory planning model based on ME-IRL.

However, simply using IRL to learn the reward function weights is not sufficient to fully express implicit intention. In our research, we first extract the priors of human drivers and utilize them to shape the expected trajectory space, thereby formulating a social trajectory set. Then, ME-IRL is employed to learn the human driving behavior and trajectory selection mechanism.

\begin{figure*}[!htbp]
  \centering
  \includegraphics[width=0.7\textwidth]{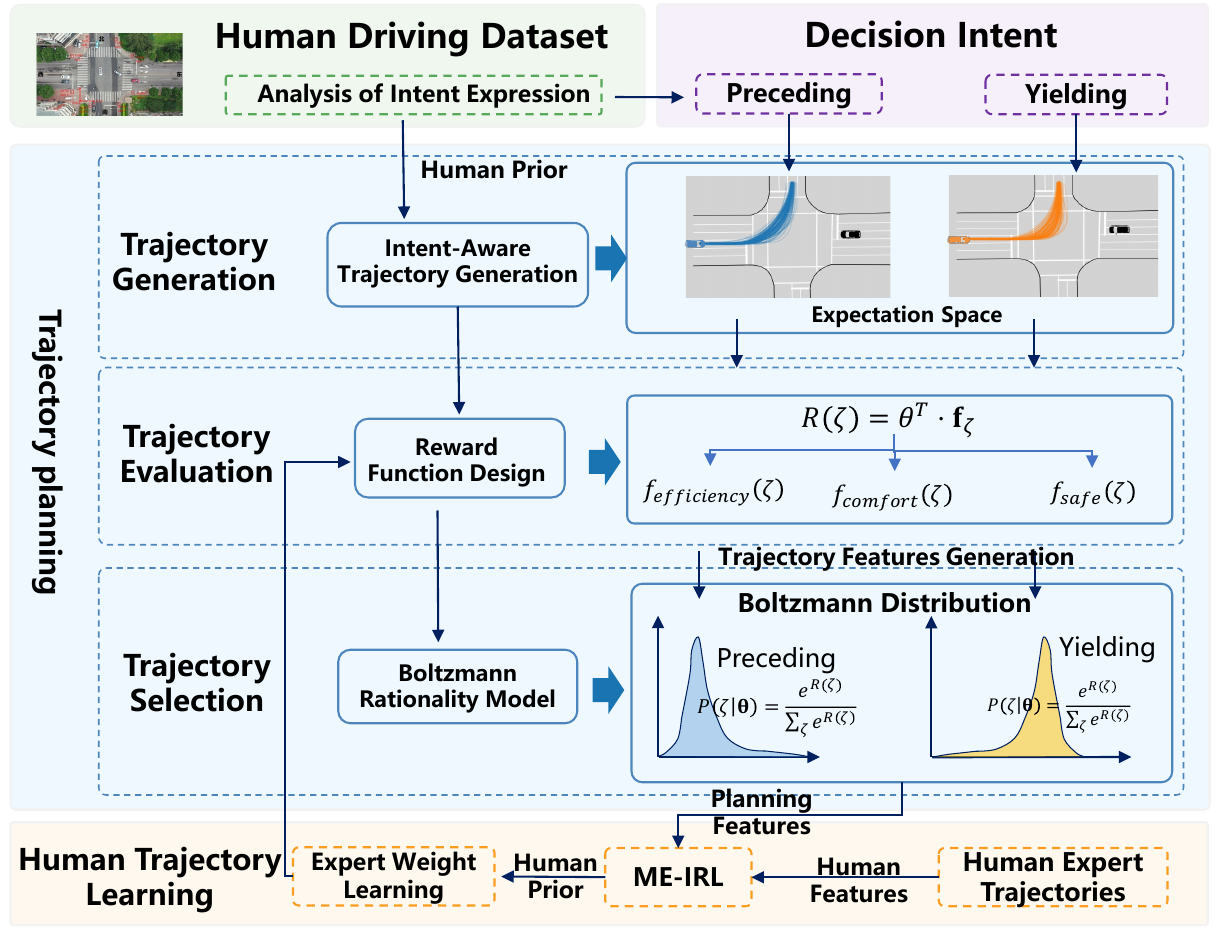}
  \caption{The comprehensive left-turn trajectory planning framework.}
  \label{fig:framework}
\end{figure*}

\section{Methodology}
\label{sec:3}
In this section, we present our framework for social trajectory planning. We begin by providing an overview of the entire framework, followed by an analysis of the left-turn trajectory characteristics of human drivers. Finally, our trajectory generation method is detailed.

\subsection{Overview of Framework}
The overall trajectory planning framework is illustrated in Fig.~\ref{fig:framework}. The specific steps of the trajectory planning method are as follows: (1) Based on the SinD dataset \cite{xu2022drone}, we analyze the intention expression characteristics of human trajectories in left-turn scenarios to provide prior knowledge from human drivers for trajectory generation; (2) Using decision intentions, such as proceeding or yielding, and combining them with the prior knowledge of trajectory intention expression, multiple candidate trajectories are generated to form the expected decision trajectory space; (3) Next, ME-IRL is employed to learn and model human driving behavior in evaluating and selecting trajectories, and design a reward function that expresses trajectory characteristics in terms of traffic efficiency, driving comfort, and interaction safety; (4) We then establish a probability distribution for the candidate trajectories based on the Boltzmann distribution, using a Boltzmann noise rationality model to simulate driver behavior in trajectory selection and complete the trajectory planning process.

After completing the algorithm training, the method will be tested and validated through simulation experiments and human-in-the-loop driving simulation experiments.


\subsection{Human Intent Expression Characteristics Analysis}
\begin{figure*}[!htbp]
    \centering
    \includegraphics[width=0.8\linewidth]{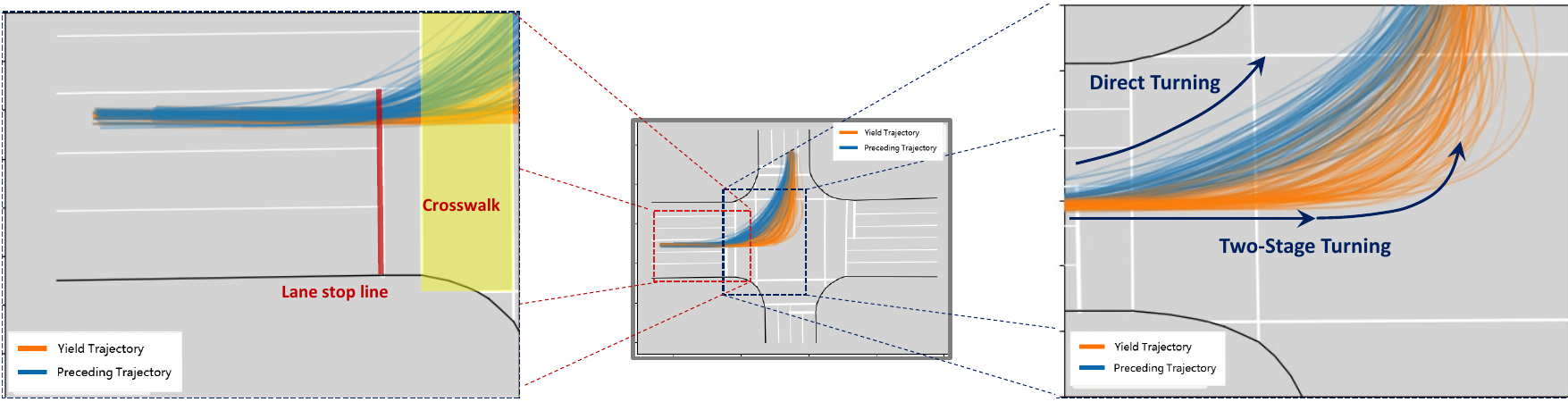}
    \caption{Trajectory characteristics under different HVs' decision-making.}
    \label{fig:Traj_HV_Prior}
\end{figure*}

Unlike existing autonomous driving algorithms, human drivers often proactively express their intentions implicitly to other drivers through speed changes and trajectory shifts in certain interaction scenarios. At intersections, where movement constraints are minimal and flexible, left-turning HVs can express their decision intentions clearly through distinct trajectory maneuvers during interactions. We use the unprotected left-turn interaction as a typical scenario to analyze the intention expression characteristics of human trajectories using the SinD dataset. The distribution of unprotected left-turn interaction trajectories in the SinD dataset is shown in Fig.~\ref{fig:Traj_HV_Prior}. The orange lines represent yield trajectories, while the blue lines represent proceed trajectories. It can be observed that human drivers adopt different intention expression methods for left-turn trajectories based on different decisions:
\begin{itemize}
    \item The proceed trajectories, as shown in Fig.~\ref{fig:Traj_HV_Prior}(a),  involve pre-turn behaviors, such as early steering and lateral displacement on the upstream segment, allowing drivers to complete the turn quickly through direct steering within the intersection.
    \item In Fig.~\ref{fig:Traj_HV_Prior}(b), yield trajectories maintain a straight path without significant displacement on the upstream segment and complete the turn within the intersection using a two-stage maneuver: first continuing straight and then making the turn.
\end{itemize}

Based on these intention expression methods corresponding to different decisions, we will establish an expected trajectory space constraint that allows for implicit intention expression, ensuring that the trajectory operates within the human-expected space and possesses the ability to convey intention.



\subsection{Left-turn Candidate Trajectories Generation}

To efficiently generate smooth and flexible trajectories at intersections, we assume that drivers typically make short-term plans based on the current interaction motion state, considering both lateral and longitudinal expected target points. Since the polynomial trajectory generation method can independently establish expressions for longitudinal and lateral displacement, and it has advantages like smooth trajectories, curvature continuity, and derivability of mathematical expressions, we employ the polynomial method to generate left-turn candidate trajectories based on the Frenet coordinate system.

We use polynomials to fit the expected trajectory, decomposing the three-dimensional optimization problem into two separate optimization problems in the longitudinal direction \(s\) and the lateral direction \(l\), represented as \(s(t)\) and \(l(t)\), respectively. The longitudinal and lateral trajectories are expressed using a quartic polynomial and a quintic polynomial, respectively, as follows:

\begin{equation}
    \begin{aligned}
    s(t) &= a_0 + a_1 t + a_2 t^2 + a_3 t^3 + a_4 t^4, \\
    l(t) &= b_0 + b_1 t + b_2 t^2 + b_3 t^3 + b_4 t^4 + b_5 t^5.
    \end{aligned}
\end{equation}

The boundary conditions for trajectory generation using the polynomial method can be represented as follows:

\begin{align}
\begin{cases}
s\left(t_0\right)=s_0, \dot{s}\left(t_0\right)=v_{s0}, \ddot{s}\left(t_0\right)=a_{s0}, \\
\dot{s}\left(T\right)=v_{sT}, \ddot{s}\left(T\right)=a_{sT}\\
l\left(t_0\right)=l_0, \dot{l}\left(t_0\right)=v_{l0}, \ddot{l}\left(t_0\right)=a_{l0}, l\left(T\right)=l_T,\\
\dot{l}\left(T\right)=v_{lT}, \ddot{l}\left(T\right)=a_{lT}
\end{cases}
\end{align}
where $s_0$, $v_{s_0}$, and $a_{s_0}$ represent the longitudinal displacement, velocity, and acceleration at the initial state,  respectively. $l_0$, $v_{l0}$, and $a_{l_0}$ denote the lateral displacement, velocity, and acceleration at the initial state. $v_{s_T}$, $a_{s_T}$ represent the longitudinal velocity and acceleration at the terminal state. Similarly, $l_T$, $v_{l_T}$, and $a_{l_T}$ denote the lateral displacement, velocity, and acceleration at the terminal state.

Meanwhile, left-turning usually requires determining a reasonable sampling space to generate a set of trajectories, from which the optimal trajectory is selected. The sampling space for generating the trajectories is determined by the range of the terminal state parameters in the boundary conditions. To avoid the dimension explosion problem caused by too many parameters, we set the terminal moment longitudinal acceleration $a_{s_T}=0$ and lateral acceleration $a_{l_T}=0$. Then, we sample the trajectories over the time length $T$. The state variables that actually affect the trajectory sampling space are as follows:

\begin{align}
\begin{cases}
v_{sT}\in\left[v_{s0}-\Delta v_s,v_{s0}+\Delta v_s\right]\\
v_{lT}\in\left[v_{l0}-\Delta v_l,v_{l0}+\Delta v_l\right]\\
l_T=f_l\left(s_T\right)\\
T=f_T\left(s_T\right)
\end{cases}
\end{align}

The change range of the longitudinal and lateral velocities, $v_{s_T}$ and $v_{l_T}$, under the trajectory terminal state is constrained based on the longitudinal and lateral velocities, $v_{s_0}$ and $v_{l_0}$, under the initial state of the trajectory. The lateral deviation $l_T$ under the terminal state is constrained based on the feature distribution of the expected trajectory space in subsequent research on human left-turn trajectory decisions. In determining the terminal position, we generate trajectories using an indefinite time length $T$. When the terminal position is uncertain, we generate trajectories using a fixed time length $T=5s$. These two methods are applied to upstream road sections and internal intersection trajectory generation, respectively.

\subsection{Desired Trajectory Space Constraint Based on Intention Expression}
\subsubsection{Constraint for Upstream Sections}
\begin{figure}[!htbp]
    \centering
    \includegraphics[width=0.9\linewidth]{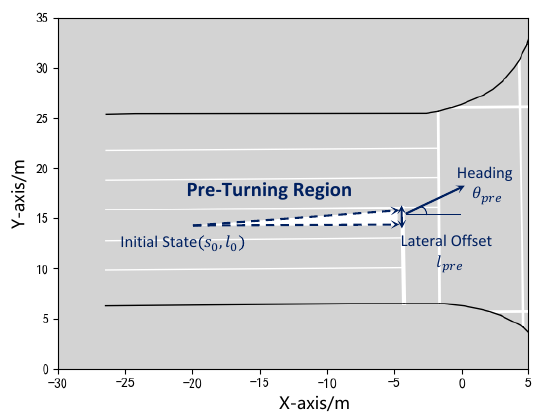}
    \caption{Space constraint of pre-steering trajectory in upstream section.}
    \label{fig:pre_turning}
\end{figure}

It was analyzed in the previous sections that for left turn trajectories in the upstream sections, the preceding trajectories tend to make a pre-turn while the yielding trajectories continue to maintain straight-line motion along the lane. As shown in Fig.~\ref{fig:pre_turning}, the position, speed, and acceleration $(s_0,l_0,v_{s0},v_{l0},a_{s0},a_{l0})$ are determined as the initial states of the scene. Based on the analysis of pre-turning behavior characteristics, it is determined that the trajectory will experience a lateral deviation and a change in heading angle when reaching the stop line. Hence, the pre-turning state of the trajectory at the stop line can be represented as $(s_{pre},l_{pre},\theta_{pre})$.
Here, $s_{pre}=s_{stop}$, the longitudinal position of the stop line on the reference line, is known. The trajectory's lateral deviation $l_{pre}$ and heading angle $\theta_{pre}$ at the stop line are determined as follows:
\begin{equation}
    \left\{\begin{matrix}\theta_0=\frac{v_{l0}}{v_{s0}}\\\theta_{pre}=\frac{v_{lpre}}{v_{spre}}\\l_{pre}\in\left[l_0,l_{stop}^{max}\right]\\\theta_{pre}\in\left[\theta_0,\theta_{stop}^{max}\right]\\\end{matrix}\right.
\end{equation}
where $l_{stop}^{max}$ is the maximum trajectory deviation at that point. $\theta_{pre}$ determines the ratio of the lateral speed $v_{lpre}$ to the longitudinal speed $v_{spre}$ of the trajectory at the stop line. $\theta_{stop}^{max}$ is the maximum heading angle of the trajectory at that point. Lateral deviation and heading angle are determined through a uniformly distributed random sampling method.

Therefore, assuming the initial trajectory state at time $(s_t,l_t,v_{st},v_{lt},a_{st},a_{lt})$, the trajectory sampling time length is set as $T_{pre}$. The pre-turning behavior characteristic state $(s_{pre},l_{pre},\theta_{pre})$ corresponds to the trajectory generation control terminal state $(s_{pre},l_{pre},v_{spre},v_{lpre},a_{spre},a_{lpre})$ at this moment, where acceleration items are simplified to default zero. Then, when controlling the trajectory generation space variable duration $T_{pre}$, longitudinal speed $v_{spre}$, and lateral speed $v_{lpre}$, they can be determined as the following calculation results:
\begin{equation}
    \left\{\begin{matrix}\mathrm{\ }\mathrm{T}_{pre}\in\left[\frac{s_{pre}-s_t}{v_{st}}-\Delta T_{pre},\frac{s_{pre}-s_t}{v_{st}}+\Delta T_{pre}\right]\\
    \sqrt{{v_{\mathrm{spre\ }}}^2+{v_{\mathrm{lpre\ }}}^2}\in\\
    \left[\sqrt{{v_{st}}^2+{v_{lt}}^2}-\Delta v_{pre},\sqrt{{v_{st}}^2+{v_{lt}}^2}+\Delta v_{pre}\right]\\\frac{v_{\mathrm{lpre\ }}}{v_{\mathrm{spre\ }}}=\theta_{pre}\\\end{matrix}\right.
\end{equation}
where $\Delta T_{pre}$ and $\Delta v_{pre}$ are control variables of trajectory duration and speed variation interval, respectively. The space variables of the yielding trajectory generation are consistent with the preceding trajectory, both controlling the trajectory sampling duration $T_{pre}$ , longitudinal speed $v_{spre}$ , and lateral speed $v_{lpre}$. The calculation method is consistent with the aforementioned one.

\subsubsection{Constraint within Intersections}
Analysis from previous sections determined that, within an intersection, left-turning vehicles following a trajectory that has the right of way will complete a rapid turn directly, while those yielding will adopt a two-phase turning method, maintaining a straight path before completing the turn.

As shown in Fig.~\ref{fig:Traj_HV_Prior}, different decision trajectories lead to distinctly different trajectory distribution spaces within the intersection due to the different turning methods employed.
In this study, we model the relationship between the longitudinal displacement, s, and lateral displacement, l, of a trajectory using a continuously varying normal distribution. The mean and standard deviation of the preemptive trajectory's normal distribution are denoted as $\mu_{\text{preempt}}(s)$ and $\sigma_{\text{preempt}}(s)$ respectively. The mean and standard deviation of the yielding trajectory's normal distribution are denoted as $\mu_{\text{yield}}(s)$ and $\sigma_{\text{yield}}(s)$ respectively.

By determining the longitudinal displacement at time t, $s(t)$, we can deduce that $l_t \sim N(\mu(s(t)),\sigma(s(t)))$. Further, based on the 3-sigma rule of normal distribution, we can determine the range of the trajectory's lateral displacement at the end, $l_t \in [\mu(s(t)) - 2\sigma(s(t)), \mu(s(t)) + 2\sigma(s(t))]$.

Simultaneously, the longitudinal speed, $v_{sT}$, and the lateral speed, $v_{lT}$, at the end of the trajectory are determined within a changing interval based on the initial speeds $v_{s0}$ and $v_{l0}$. These speeds are uniformly sampled within the interval to determine $v_{sT}$ and $v_{lT}$.
Assuming an initial trajectory state within the intersection at a particular time, $(s_0, l_0, v_{s0}, v_{l0}, a_{s0}, a_{l0})$, the longitudinal displacement at the end of the trajectory, $s_T$, can be computed given $v_{sT}$. With $s_T$ determined, the lateral displacement at the end of the trajectory, $l_T$, can be decided based on the established relationship between $s$ and $l$.

Therefore, the expected trajectory space within the intersection can be controlled and generated by the following variables:
\begin{equation}
    \left\{
\begin{matrix}
v_{sT} \in [v_{s0} - \Delta v_s, v_{s0} + \Delta v_s]\\
v_{lT} \in [v_{l0} - \Delta v_l, v_{l0} + \Delta v_l]\\
l_T \in [\mu(s_T) - 2\sigma(s_T), \mu(s_T) + 2\sigma(s_T)]
\end{matrix}
\right.
\end{equation}
where $\Delta v_s$ and $\Delta v_l$ are the control variables for the changing interval of longitudinal and lateral speeds respectively.

In order to ensure that all generated trajectories reflect normal kinematic characteristics exhibited by human driving and avoid collisions, we apply the following constraints to the dynamics of the trajectory:

\begin{equation}
    \left\{
\begin{matrix}
v_{min} \le v_t \le v_{max}\\
a_{min} \le a_t \le a_{max}\\
c_{min} \le c_t \le c_{max}
\end{matrix}
\right.
\end{equation}
where $v_t$, $a_t$, and $c_t$ represent the speed, acceleration, and curvature of a generated trajectory $\zeta_I^i \in \zeta_I$ at time $t \in [0, T]$. The $(v_{\text{max}}, v_{\text{min}})$, $(a_{\text{max}}, a_{\text{min}})$, and $(c_{\text{max}}, c_{\text{min}})$ are the minimum and maximum speeds, accelerations, and curvatures respectively, as determined from the statistical analysis of real human trajectories.

Furthermore, we conduct collision checks to ensure that the generated trajectories do not result in collisions with interactive objects at any time. This study includes the consideration of a safety margin, determining the safe bounding box $Boundingbox_{\text{safe}}$. The collision constraint can be expressed as:

\begin{equation}
Boundingbox_{\text{safe}}^t\left(\text{left}\right) \cap Boundingbox_{\text{safe}}^t\left(\text{straight}\right) = \emptyset \quad 
\end{equation}
where $Boundingbox_{\text{safe}}^t$ is the safe bounding box at time t, which is calculated by adding 0.5m to the front and rear spaces and 0.3m to the left and right spaces of the vehicle's boundary rectangle. If, at any point, a trajectory $\zeta_{I,K}^i \in \zeta_{I,K}$ fails to meet the collision constraint, that trajectory is removed from the set $\zeta_{I,K}$. The final set of desired trajectories that meet the collision constraints is $\zeta_{I,K,C}$.

\subsection{Human Prior Learning}
In traditional trajectory planning research, after defining the trajectory feature function, the optimal trajectory selection is achieved by manually setting or experimentally determining the trajectory reward function feature weights. However, in real-world interaction scenarios, it's challenging to accurately specify a reward function that captures all aspects of safe and efficient driving. IRL can solve this problem. In this subsection, the reward function is first designed and ME-IRL is utilized to learn the human expert's trajectory behavior selection strategy under different decisions.

During learning process, we assume that the total reward of a trajectory is a linear expression of the trajectory reward function, which is the weighted sum of selected features. Furthermore, we postulate that human drivers' preferences or behaviors under the same decision do not exhibit noticeable time variability and individual heterogeneity. Therefore, the total reward $R\left(\zeta\right)$ of a trajectory $\zeta$ can be expressed as:
\begin{equation}
R\left(\zeta\right)=\theta^T\cdot\mathbf{f}\zeta
\end{equation}
where $\theta=\left[\theta_1,\theta_2,\cdots,\theta_K\right]$ is the weight vector, $\mathbf{f}\zeta$ is the reward function vector of trajectory $\zeta$, and $K$ depends on the number of trajectory reward functions $\mathbf{f}\zeta$.

\subsubsection{Reward Function Design}
When designing the reward function, we considered three aspects: traffic efficiency, driving comfort, and interactive safety.

\textbf{Traffic Efficiency.}
We set the reward function for traffic efficiency as the loss in speed, which is the difference between the trajectory speed and the expected speed. We determine the traffic efficiency feature $f_{\text{efficiency}}\left(\zeta\right)$ of the trajectory $\zeta$ to be the mean speed loss at all times, represented as follows:
\begin{equation}
f_{\text{efficiency}}\left(\zeta\right)=-\frac{\sqrt{\sum_{t=0}^{T}\left(v_t-v_{\text{target}}\right)^2}}{T}
\end{equation}
where $v_t$ is the scalar speed of the trajectory $\zeta$ at time $t$, $T$ is the total duration of trajectory sampling, and $v_{\text{target}}$ is the expected speed of the vehicle turning left.

\textbf{Driving Comfort.}
We select the jerk (rate of change of acceleration) to establish the comfort feature function $f_{\text{comfort}}\left(\zeta\right)$ for assessing whether the trajectory $\zeta$ is comfortable. This feature is determined as the mean of the jerk vector sum longitudinally and laterally at all times for the trajectory $\zeta$, calculated as follows:

\begin{equation}
    \left\{\begin{matrix}
    f_{\text{comfort}}\left(\zeta\right)= - \frac{\sum_{t=0}^{T}\sqrt{((Jerk_s^t)^2 + (Jerk_l^t)^2)}}{T}\\
    Jerk_s^t = s'''(t)\\
    Jerk_l^t = l'''(t)\\
    \end{matrix}
    \right.
\end{equation}
where $Jerk_s^t$ and $Jerk_l^t$ are the longitudinal and lateral jerks of the trajectory $\zeta$ at time $t$ respectively. The reward function $f_{\text{comfort}}\left(\zeta\right)$ takes into account the smoothness in both the longitudinal and lateral directions.

\textbf{Interaction Safety.}
We quantify interaction safety by calculating the time difference $\Delta TTCP$ between the interacting parties reaching the conflict point. $\Delta TTCP$ takes into account the relative relationships of the positions and speeds of the interacting parties, characterizing the time interval for both parties to leave the conflict point under the current state.

The interaction safety is deconstructed into longitudinal progress and lateral deviation. The impact on interaction safety from the longitudinal progress at time $t$ is determined as the time difference for the interacting parties to reach the conflict point without considering the lateral deviation of the left-turning vehicle, denoted as $\Delta TTCP_{st}$. Its calculation is as follows:

\begin{equation}
\Delta TTCP_{st}=TTCP_{st}^{left}-TTCP_{st}^{straight}
\end{equation}
\begin{align}
\begin{cases}
TTCP_{st}^{left}=\frac{s_t^{left}-s_{cp}^{left}}{v_{st}^{left}}\\
TTCP_{st}^{straight}=\frac{s_t^{straight}-s_{cp}^{straight}}{v_{st}^{straight}}
\end{cases}
\end{align}
where $TTCP_{st}^{left}$ and $TTCP_{st}^{straight}$ are the times for the left-turning and straight-driving vehicles to pass the conflict point at the longitudinal level at time $t$ respectively.
$s_t^{left}$ and $s_t^{straight}$ are the longitudinal positions of the left-turning and straight-driving vehicles at time $t$, $s_{cp}^{left}$ and $s_{cp}^{straight}$ are the longitudinal positions of the reference conflict points on the reference lines for the left-turning and straight-driving vehicles respectively, $v_{st}^{left}$ and $v_{st}^{straight}$ are the longitudinal velocities of the left-turning and straight-driving vehicles at time $t$.

\begin{figure}[!htbp]
    \centering
    \includegraphics[width=0.9\linewidth]{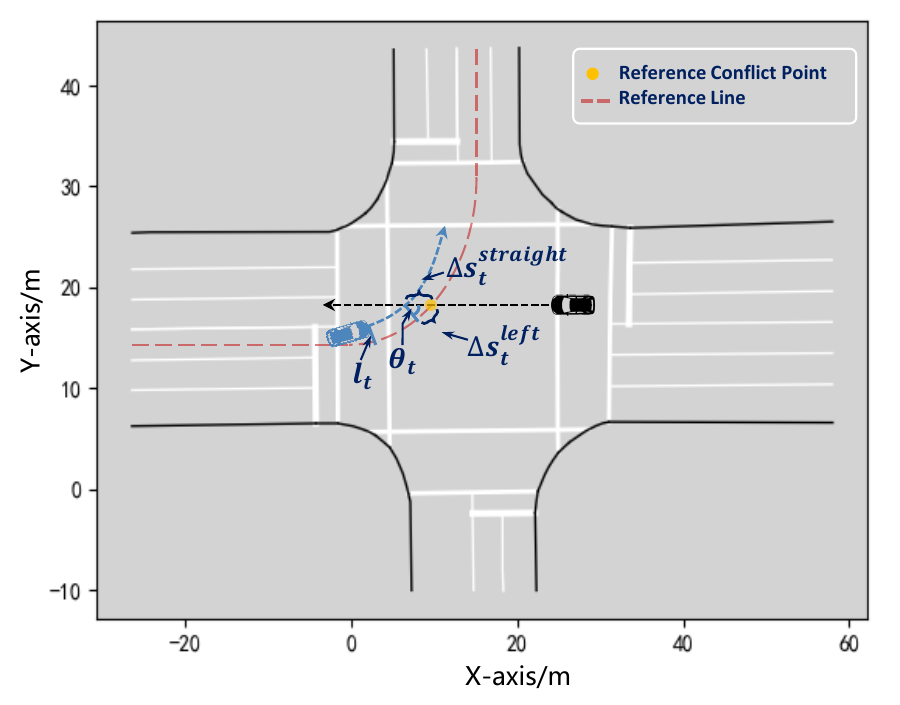}
    \caption{Influence of lateral displacement of left-turning vehicle on interactive safety.}
    \label{fig:IRL_Safety_Term}
\end{figure}

After calculating $\Delta TTCP_{st}$, we further determine the feature function $f_{safe,s}\left(\zeta\right)$ on the longitudinal level of interaction safety as the mean of $\Delta TTCP_{st}$ at all times on trajectory $\zeta$, represented as follows:
\begin{equation}
f_{safe,s}\left(\zeta\right)=\frac{\sum_{t=0}^{T}\left|\Delta TTCP_{st}\right|}{T}
\end{equation}

On the other hand, the impact of lateral deviation at time $t$ on interaction safety, as analyzed above, is determined by the impact of the left-turning vehicle's lateral deviation on the time difference for both interacting parties to reach the conflict point, denoted as $\Delta TTCP_{lt}$. Its calculation is as follows:

\begin{equation}
\Delta TTCP_{lt}=\Delta TTCP_l^{left}+\Delta TTCP_l^{straight}
\end{equation}
where $\Delta TTCP_l^{left}$ and $\Delta TTCP_l^{straight}$ are the impacts caused by the left-turn vehicle's lateral deviation on the times for itself and the oncoming straight-driving vehicle to pass through the conflict point, respectively. Their calculations consider the variables as shown in Fig.~\ref{fig:IRL_Safety_Term} and are as follows:

\begin{align}
\begin{cases}
\Delta TTCP_l^{left}=\frac{l_t \cdot tan{\theta_t}}{v_{st}^{left}}\\
\Delta TTCP_l^{straight}=\frac{l_t \cdot cos{\theta_t}}{v_{st}^{straight}}
\end{cases}
\end{align}
where $l_t$ is the lateral deviation of the left-turn vehicle at time $t$, and $\theta_t$ is the angle produced by the expected trajectory of the oncoming straight-driving vehicle and the projection point of the expected conflict point on the reference line along the $l$ axis at time $t$.

After calculating $\Delta TTCP_{lt}$, we further determine the feature function $f_{safe,l}\left(\zeta\right)$ on the longitudinal level of interaction safety as the mean of $\Delta TTCP_{lt}$ at all times on trajectory $\zeta$, represented as follows:

\begin{equation}
f_{safe,l}\left(\zeta\right)=\frac{\sum_{t=0}^{T}\left|\Delta TTCP_{lt}\right|}{T}
\end{equation}

The interaction safety feature function of trajectory $\zeta$ is characterized from both longitudinal and lateral perspectives, represented as $f_{safe,s}\left(\zeta\right)$ and $f_{safe,l}\left(\zeta\right)$.

\subsubsection{Maximum Entropy IRL Learning}

Given a human driving demonstration dataset $D=\{ \zeta_1,\zeta_2,\cdots,\zeta_N \}$ composed of $N$ trajectories, the ME-IRL algorithm is used to infer the reward weight $\theta$, which can then be used to generate driving strategies that match the human expert demonstration trajectories.

Simultaneously, to simulate the randomness of human driver's trajectory selection, we employ the Boltzmann noise theory model to construct the candidate trajectory distribution. Under the Boltzmann distribution, all features expected to match the expert's demonstration have the maximum entropy principle, corresponding to the ME-IRL. Therefore, the probability of a trajectory is proportional to the return of that trajectory,
\begin{equation}
P\left(\zeta\left|\mathbf{\theta}\right.\right)=\frac{e^{R\left(\zeta\right)}}{Z\left(\theta\right)}=\frac{e^{\mathbf{\theta}^T\mathbf{f}\zeta}}{Z\left(\theta\right)}
\end{equation}
where $P\left(\zeta\left|\mathbf{\theta}\right.\right)$ is the probability of trajectory $\zeta$, and $Z\left(\theta\right)$ is the partition function. As the partition function $Z\left(\theta\right)$ represents the integral sum of the rewards of all possible trajectories, it is challenging to directly calculate in continuous and high-dimensional spaces. We reduce the trajectory space to the previously researched and mined human prior expected trajectory space $\zeta_{I,K,C}$ and use a finite number of discretely generated feasible trajectories to approximate the partition function. Therefore, the probability expression of a trajectory yields that
\begin{equation}
P\left(\zeta\left|\mathbf{\theta}\right.\right)\approx\frac{e^{\mathbf{\theta}^T\mathbf{f}\zeta}}{\sum_{i=1}^{N}e^{\mathbf{\theta}^T\mathbf{f}{{\widetilde{\zeta}}^i}}}
\end{equation}
where ${\widetilde{\zeta}}^i\in\zeta{I,K,C}$ is a generated trajectory with the same initial state as trajectory $\zeta$, $\mathbf{f}_{{\widetilde{\zeta}}^i}$ is the feature vector of trajectory ${\widetilde{\zeta}}^i$, and $N$ is the total number of generated trajectories.

The aim of ME-IRL is to maximize the likelihood of expert demonstration trajectories by adjusting the feature weight $\theta$. The optimization objective function can be expressed as:
\begin{equation}
\max_{\theta} \mathcal{J}(\mathbf{\theta})= \max_{\theta} 
\sum_{\zeta \in \mathcal{D}} 
 \log P(\zeta | \theta)
\end{equation}
where $\mathcal{D}=\{ \zeta_i \}{i=1}^N$ represents the set of human expert demonstration trajectories. By substituting $P\left(\zeta\left|\mathbf{\theta}\right.\right)$ into the above equation, we obtain the optimized objective function $\mathcal{J}\left(\theta\right)$ as follows:
\begin{equation}
\mathcal{J}(\mathbf{\theta})=\sum_{\zeta\in\mathcal{D}}\left[\mathbf{\theta}^T\mathbf{f}_\zeta-log{\sum_{i=1}^{M}e^{\mathbf{\theta}^T\mathbf{f}_{{\widetilde{\zeta}}^i}}} \right]
\end{equation}

The above equation can be optimized using gradient-based methods. 
The gradient of the optimization objective function $\mathcal{J}\left(\theta\right)$, $\nabla\theta\mathcal{J}\left(\theta\right)$, can be expressed as follows:
\begin{equation}
\nabla_\mathbf{\theta}\mathcal{J}(\mathbf{\theta})=\sum_{\zeta\in\mathcal{D}}\left[\mathbf{f}_\zeta-\sum_{i=1}^{M}\frac{e^{\mathbf{\theta}^T\mathbf{f}_{{\widetilde{\zeta}}^i}}}{\sum_{i=1}^{M}e^{\mathbf{\theta}^T\mathbf{f}_{{\widetilde{\zeta}}^i}}}\mathbf{f}_{{\widetilde{\zeta}}^i}\right]
\end{equation}
The gradient can be viewed as the difference in feature expectations between human demonstration trajectories and generated trajectories:
\begin{equation}
\nabla_\mathbf{\theta}\mathcal{J}(\mathbf{\theta})=\sum_{\zeta\in\mathcal{D}}\left[\mathbf{f}_\zeta-\sum_{i=1}^{M}P\left({\widetilde{\zeta}}^i\left|\mathbf{\theta}\right.\right)\mathbf{f}_{{\widetilde{\zeta}}^i}\right]
\end{equation}

Following the process outlined by  \cite{huang2021driving}, we use a gradient ascent method to iteratively update the trajectory feature weights and compute the optimization objective until the loss converges. To prevent overfitting, we incorporate L2 regularization into the objective function $\mathcal{J}\left(\theta\right)$. Consequently, the gradient $\nabla\theta\mathcal{J}\left(\theta\right)$ includes the difference in feature expectations plus a regularization term, as shown below:
\begin{equation}
\nabla_\mathbf{\theta}\mathcal{J}(\mathbf{\theta})=\sum_{\zeta\in\mathcal{D}}\left[\mathbf{f}_\zeta-\sum_{i=1}^{M}P\left({\widetilde{\zeta}}^i\left|\mathbf{\theta}\right.\right)\mathbf{f}_{{\widetilde{\zeta}}^i}\right]-2\lambda\theta
\end{equation}
where $\lambda > 0$ is the regularization parameter.

The whole process is summarized in Algorithm \ref{algo:irl_weight_learning}. After learning, the learned reward function weight $\theta$ will be employed to assign scores to the candidate trajectories. The trajectory with the highest score will then be selected as the planned output.

\begin{algorithm}
\SetAlFnt{\small}
\SetKwInOut{Parameter}{Inputs}
\SetKwInOut{Output}{Output}
\caption{ME-IRL Trajectory Weight Learning Process}
\label{algo:irl_weight_learning}
\LinesNumbered 
\SetAlgoLined
\Parameter{Left-turn trajectory set \(D = \{\zeta_1, \zeta_2, \dots, \zeta_N\}\), going straight trajectory set \(P\), learning rate \(\alpha\), regularization parameter \(\lambda\), number of iterations \(E\)}
\Output{Reward function weights \(\theta^*\)}
\vspace{0.2em}
\hrule
\vspace{0.2em}

Initialize reward function weights \(\theta\) randomly;\\
Compute human trajectory features \(\bar{f} = \sum_{i=1}^{N} f(\zeta_i)\);\\
Define sampling space control variables \(\{v_{sT}, v_{lT}, l_T\}\), determine the trajectory planning time window \(T\), and generate the trajectory set \(\tilde{D}_i = \{\tilde{\zeta}_i^j\}\) with the same initial state as \(\zeta_i\);\\
Compute trajectory features \(f(\tilde{\zeta}_i^j)\) for all generated trajectories interacting with the real oncoming straight trajectory set \(P\);\\
\Repeat{iteration count \(E\) is reached}{
    Compute the feature expectation \(\tilde{f}\) for the generated trajectory samples;\\
    Calculate the gradient \(\nabla_\theta J(\theta) = \bar{f} - \tilde{f} - 2\lambda\theta\);\\
    Update the reward weights \(\theta = \theta + \nabla_\theta J(\theta)\);\\
}

\end{algorithm}

\section{Experiments on the Simulation Platform}
\label{sec:4}
In this section, we detail the dataset used and the implementation of our experiment, followed by an analytical review of the experiments conducted.

\subsection{Dataset}
Our analysis utilizes the SIND dataset \cite{xu2022drone}, a drone-sourced dataset, to study human driver interaction at intersections. Compiled by Tsinghua University's SOTIF research team, the SIND dataset comprises intersection data from Tianjin, China, characterized by a two-phase traffic signal system. This system, allowing simultaneous movements of left-turning and straight-moving vehicles, results in significant interactions and conflicts. We meticulously selected 268 one-on-one interaction events between left-turning and straight-moving vehicles from this dataset, including 132 yielding and 136 proceeding left-turning instances. More description can be found here\footnote{See \url{https://drive.google.com/file/d/1FChn8wm8rZQ-OzOUCdbmSBGx41OtdI7_/view?usp=drive_link}}.

\subsection{Implementation Details}
For inverse reinforcement learning, we first extracted expert human left-turning trajectories from the SIND dataset, lasting between 8 to 15 seconds. From the onset of the left-turning and straight-moving interaction, we selected a 5-second trajectory segment at 0.5-second intervals as the expert demonstration, denoted as $\zeta$. Trajectories incorporating intent expression, denoted as $\zeta_{I,K,C}$, are generated from these segments. We use 80\% of these generated trajectories and their corresponding expert demonstrations for training the reward function, reserving the remaining 20\% for testing.

The trajectory generation sample space is defined by three control variables: end-time longitudinal speed $v_{sT}$, lateral speed $v_{lT}$, and lateral displacement $l_T$. We uniformly selected values for $v_{sT}$ and $v_{lT}$ within their respective ranges, and 10 values for $l_T$ within its range. The planning duration is set to 5 seconds, with a 0.1-second interval between trajectory points. For the ME-IRL learning process, we set the number of iterations E to 1000, with parameters $\alpha = 0.05$ and $\lambda = 0.01$.

For comparative analysis, we chose three other trajectory planning methods: Frenet, decision-based ME-IRL without intent expression space consideration, and potential field-based motion planning. These methods were compared using real interactive event decisions as high-level decision inputs. Further details and results are available here.\footnote{See \url{https://drive.google.com/drive/folders/1hMP1QvL8jq-jti_rTme5aG1GQ7POm6K1?usp=drive_link}}

\begin{figure}[htbp]
  \centering
  \includegraphics[width=0.4\textwidth]{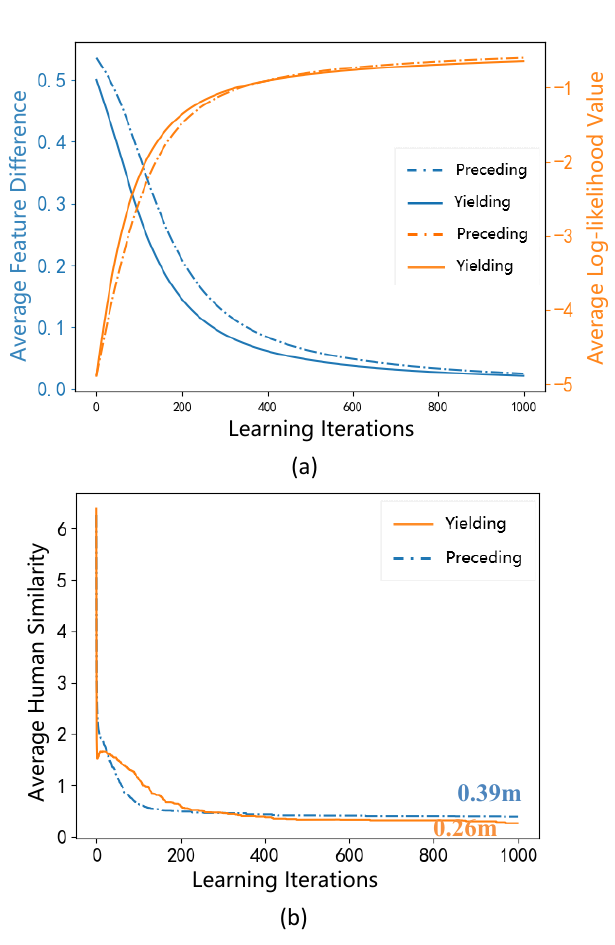}
  \caption{The learning results of IRL, (a) Convergence Verification of IRL Results, (b) Accuracy Verification of IRL. }
  \label{fig:IRL_Result}
\end{figure}

\subsection{Analysis of Inverse Reinforcement Learning Results}
The Maximum Entropy IRL model's training progression is shown in Fig.~\ref{fig:IRL_Result}. Panel (a) of the figure demonstrates the iterative process, highlighting the average feature discrepancy and the change in average log likelihood value of the human expert demonstration trajectory under yield and priority decisions. These correspond to the gradient $\nabla_\theta\mathcal{J}\left(\theta\right)$ and the optimization target function $\mathcal{J}\left(\theta\right)$. The figure shows a steady increase in the average log likelihood value, indicating convergence over iterations for different decision trajectories.

We assess the trajectory planning's closeness to human driving using Average Human Trajectory Similarity (AHL). AHL is defined as the average final displacement error of the n highest probability trajectories in the generated distribution:
\begin{equation}
\text{AHL} = \frac{1}{n} \min_{i=1}^{n} |\hat{\zeta}_i(T) - \zeta(T)|_2
\end{equation}
Here, $\hat{\zeta}_i(T)$ are the top n trajectories in the generated distribution, and $\zeta(T)$ is the actual human driver trajectory. Fig.~\ref{fig:IRL_Result} (b) depicts the ME-IRL accuracy for both preceding and yielding decisions. The priority and yield decision trajectories exhibit average errors of only 0.39m and 0.26m, respectively, compared to actual human trajectories, validating the efficacy of our method.

\subsection{Evaluation of Planning Results}
Our evaluation of the proposed trajectory planning method encompasses four dimensions: trajectory distribution, intent expression capability, safety and efficiency of interactions, and computational efficiency along with learning efficacy.

\subsubsection{Overall Trajectory Distribution}
We assessed the social aptitude of trajectory planning through its overall distribution and spatial extent. Methods with limited sociality, showing minimal consideration for intent expression, often result in a restricted range of trajectory choices. This limitation hinders the clear differentiation of trajectories under various decision-making scenarios, thus impeding intent recognition by other vehicles.

\begin{figure*} 
    \centering
    \includegraphics[width=0.7\linewidth]{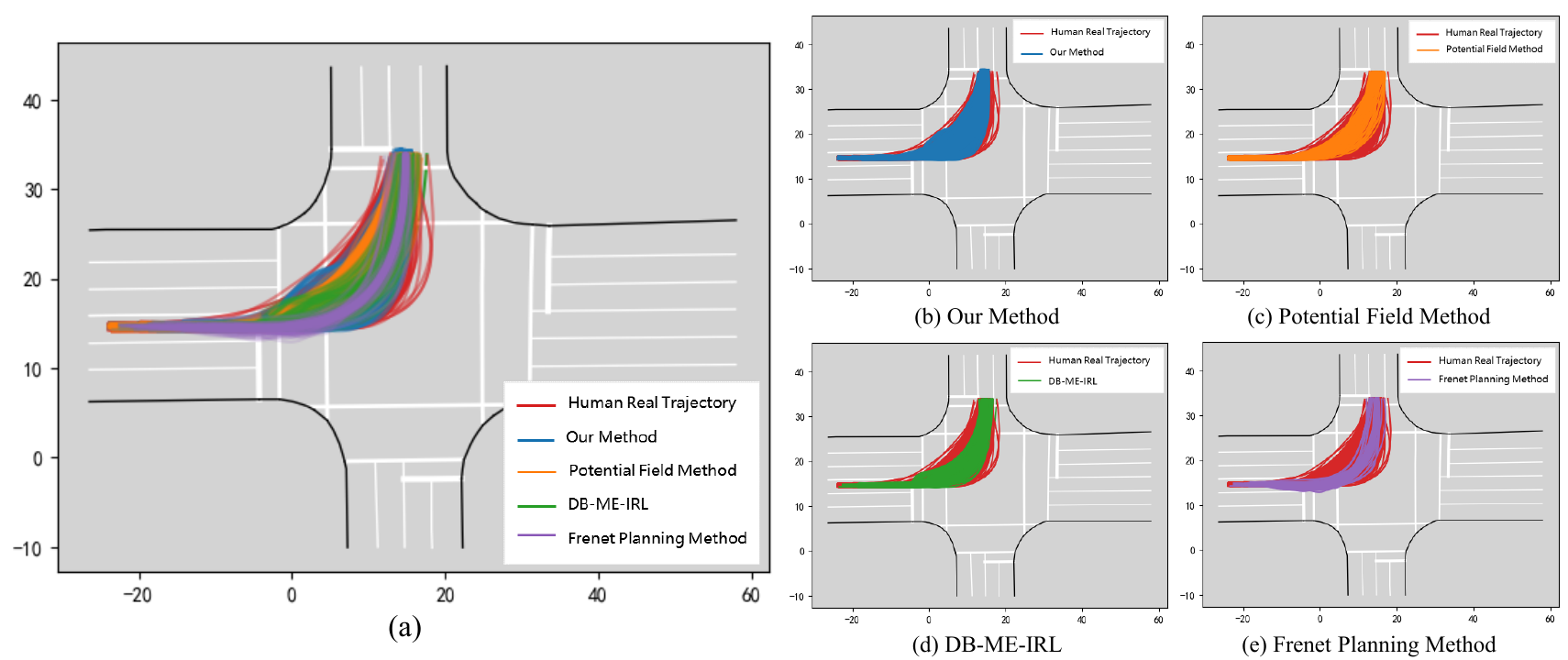}
    \caption{The planning trajectories from different methods. (a): Comparison of four trajectory methods with real trajectory distribution; (b) Our method; (c) Potential field method; (d) DB-ME-IRL; (e)Frenet planning method.}
    \label{fig:planning_results_compare}
\end{figure*}

To visually and quantitatively compare the trajectory spaces of different methods, we juxtaposed the distributions of four planning methods with actual human trajectory data, as shown in  Fig.~\ref{fig:planning_results_compare}). Qualitative analysis reveals our method's trajectory space closely aligns with actual human trajectories, while the potential field and DB-ME-IRL methods are somewhat less accurate, and the Frenet method displays the most significant deviation.

For a quantitative comparison, we employed trajectory coverage analysis. This involved dividing the upstream lane and intersection area into 0.5m subregions and checking for trajectory presence within these areas. Our method achieved a 77\% match with actual human trajectory coverage, significantly outperforming the other methods, with the Frenet method only covering 54\% of the actual trajectories.
\begin{table*} 
    \centering
    \caption{Comparison of trajectory coverage of different methods.}
    \label{table:coverage compare}
    \resizebox{\textwidth}{!}{
    \begin{tabular}{*{6}{c}}
    \toprule
        \multirow{2}{*}{Distribution characteristic index} & \multicolumn{5}{c}{Trajectory planning method}\\ 
        \cmidrule(lr){2-6}
        {} & Real trajectory & Our method & Frenet Planning Method & DB-ME-IRL & Potential field method \\
    \midrule
        Meta area coverage & 1066 & 824 & 576 & 710 & 737 \\ \hline
        Ratio to true trajectory & - & 77\% & 54\% & 67\% & 69\% \\
    \bottomrule
    \end{tabular}
    }
\end{table*}

\subsubsection{Capability of Intent Expression}
In interactions with oncoming vehicles, left-turning vehicles typically employ lateral offset as a means to convey decision intent. The capability of a trajectory plan to express intent is thus evaluated by examining the extent of lateral offset.

\begin{figure*}
    \centering
    \includegraphics[width=0.7\linewidth]{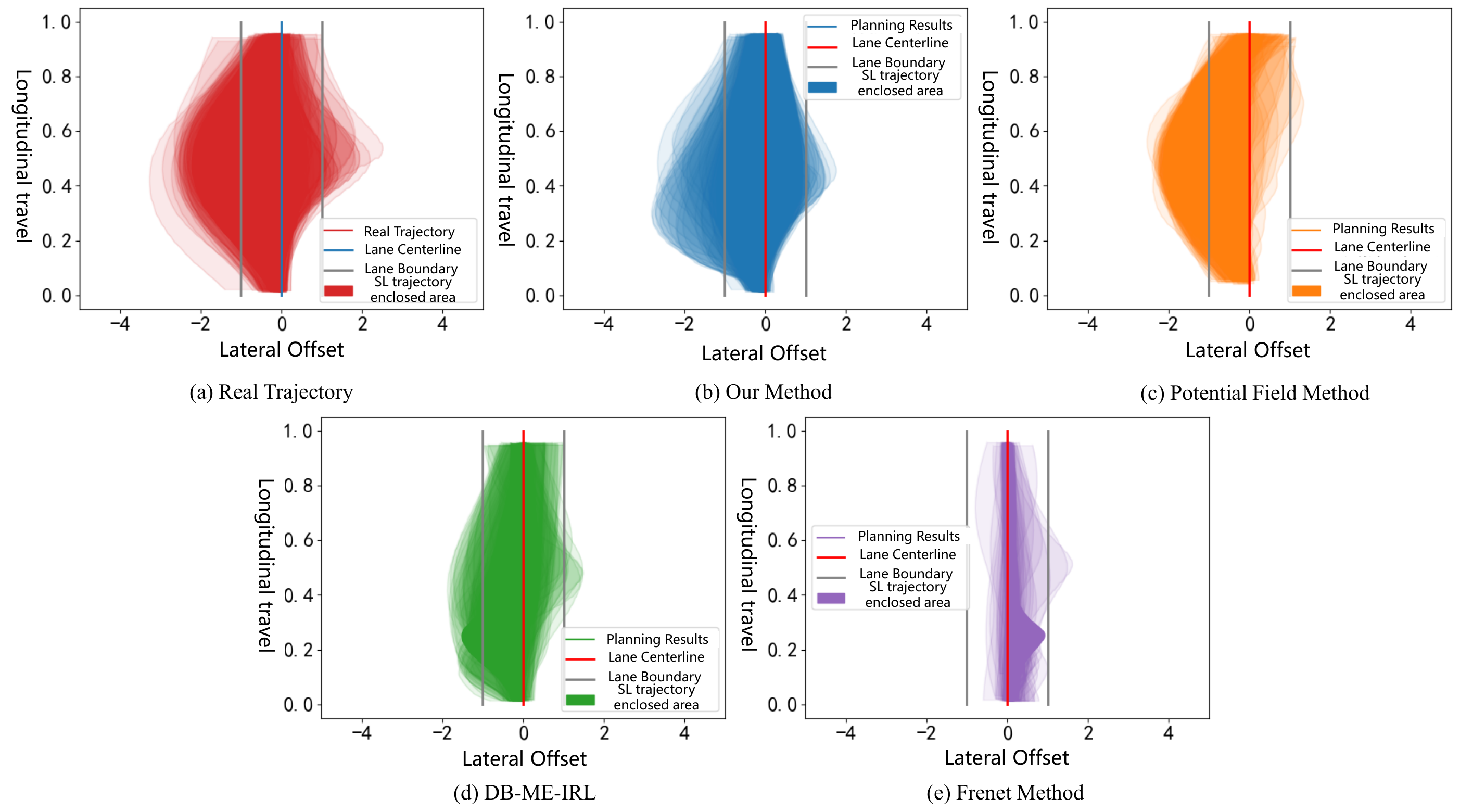}
    \caption{SL trajectory distribution under different planning methods.}
    \label{fig:SL_traj_distribution}
\end{figure*}

Fig.\ref{fig:SL_traj_distribution} displays the normalized Straight Line (SL) trajectories under various planning methods. The red (blue) line indicates the center of the virtual left-turn lane, with gray lines marking its boundaries. Offsets to the left are negative, while those to the right are positive. Our method's SL trajectory offset closely mirrors actual driver behavior, outperforming other methods which display notable deviations.

The potential field method, although capable of mimicking priority trajectories with significant leftward offsets, fails to replicate the longer straight-line path indicative of yielding behavior seen in human trajectories. The DB-ME-IRL and Frenet methods show less proficiency in intent expression. 

\begin{figure}
    \centering
    \includegraphics[width=1\linewidth]{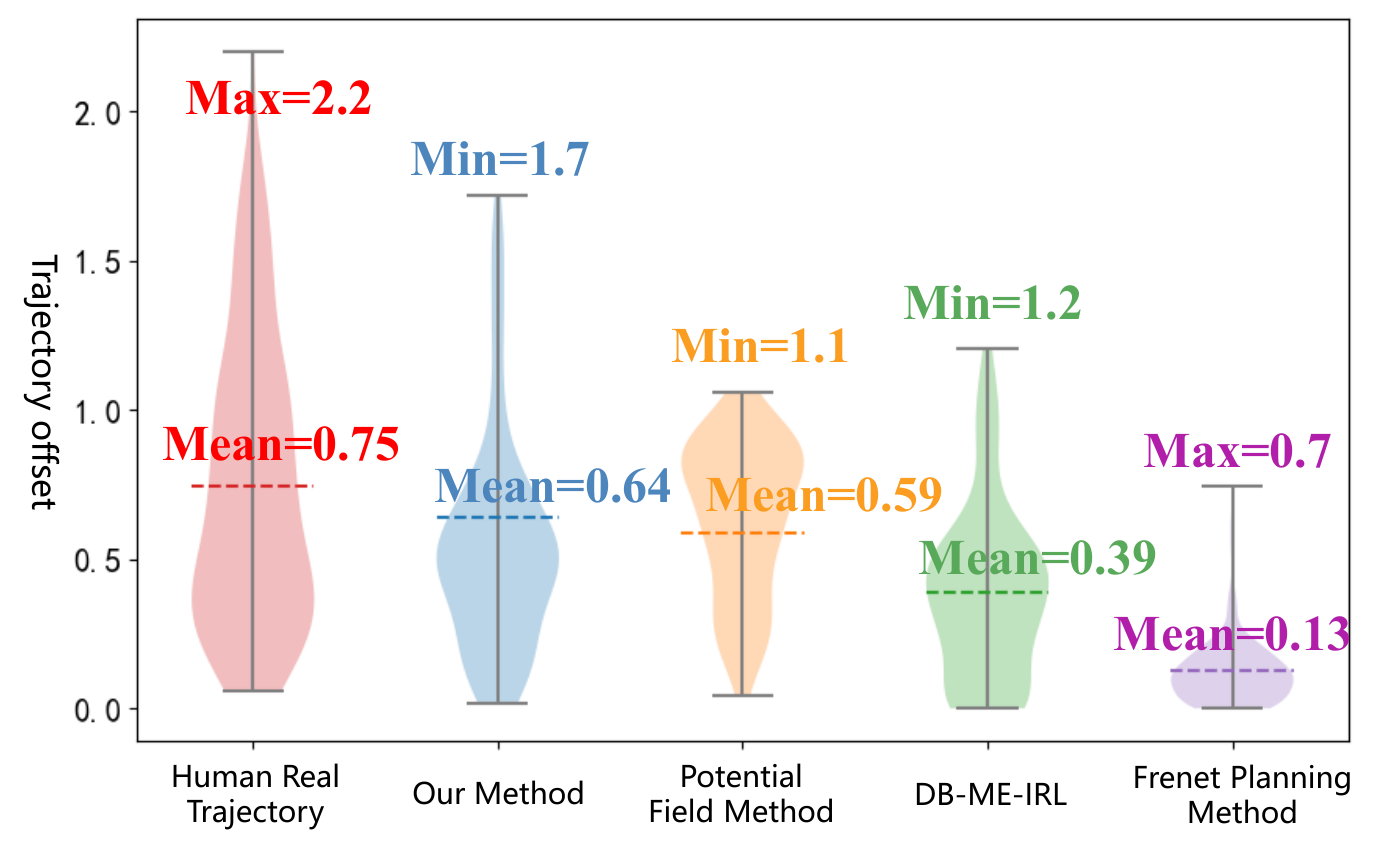}
    \caption{Total amount of normalized SL trajectory offset from real trajectory and different methods.}
    \label{fig:SL_Trajectory_Offset}
\end{figure}

The total lateral offset of a trajectory, represented by the enclosed area $S_{SL}$ between the normalized SL trajectory and the virtual lane centerline, was calculated and visualized in a violin plot (Fig. \ref{fig:SL_Trajectory_Offset}). Our method achieved an average offset of 0.64, closely following the real trajectory average of 0.75 and showing an 8\% improvement over comparative methods. In terms of maximum offset, our method reached 1.7 compared to the real trajectory's 2.2, marking a 42\% improvement over other methods.

\subsubsection{Evaluation of Motion Interaction Features}
The safety and efficiency of the trajectory planning methods were assessed using Post-Encroachment Time (PET) and Travel Time metrics.

\begin{figure}
    \centering
    \includegraphics[width=1\linewidth]{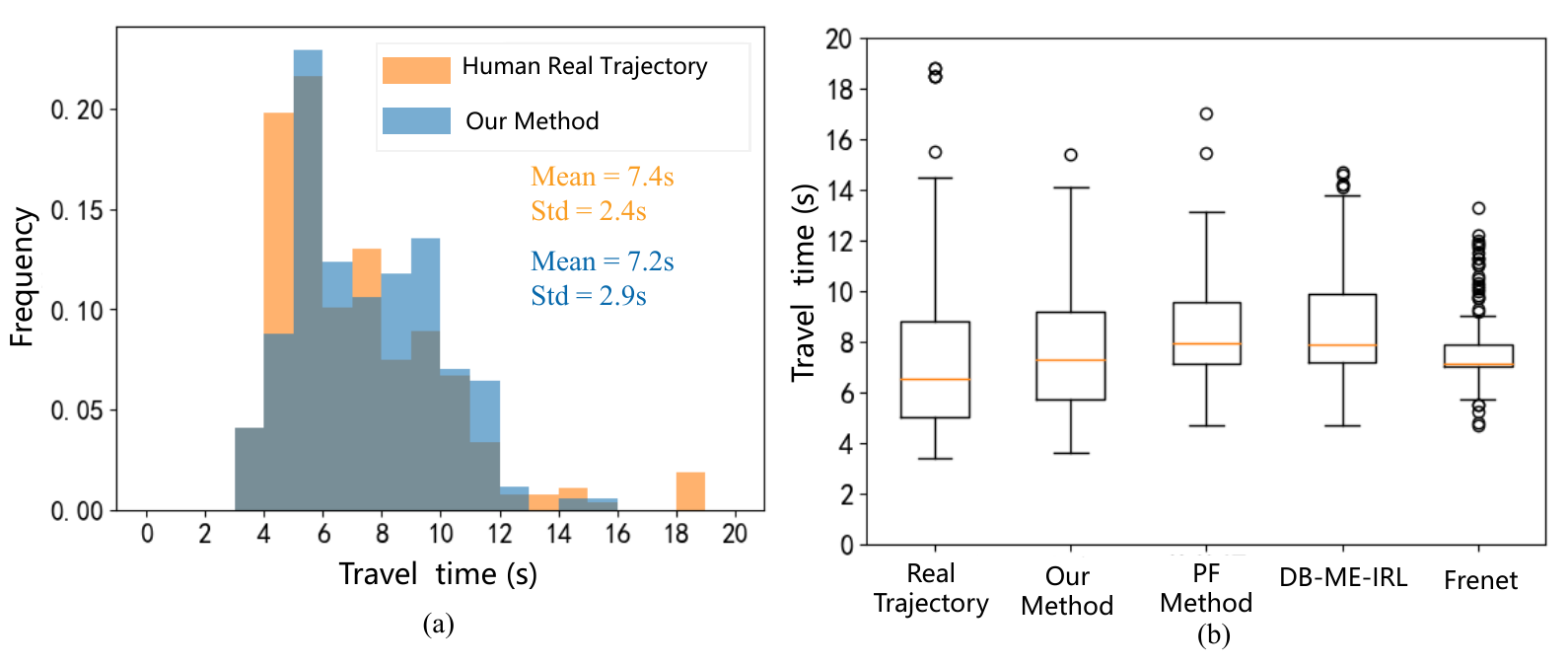}
    \caption{Comparison of travel time distribution of planning trajectories in intersections.}
    \label{fig:travel_time_distri}
\end{figure}

Travel time within the intersection was analyzed (Fig. \ref{fig:travel_time_distri}). Our method's average travel time of 7.4s with a standard deviation of 2.4s closely approximates the real trajectory average of 7.2s. Comparative analysis across different methods indicated our method aligns most closely with real trajectory travel times.

\begin{figure}
    \centering
    \includegraphics[width=1\linewidth]{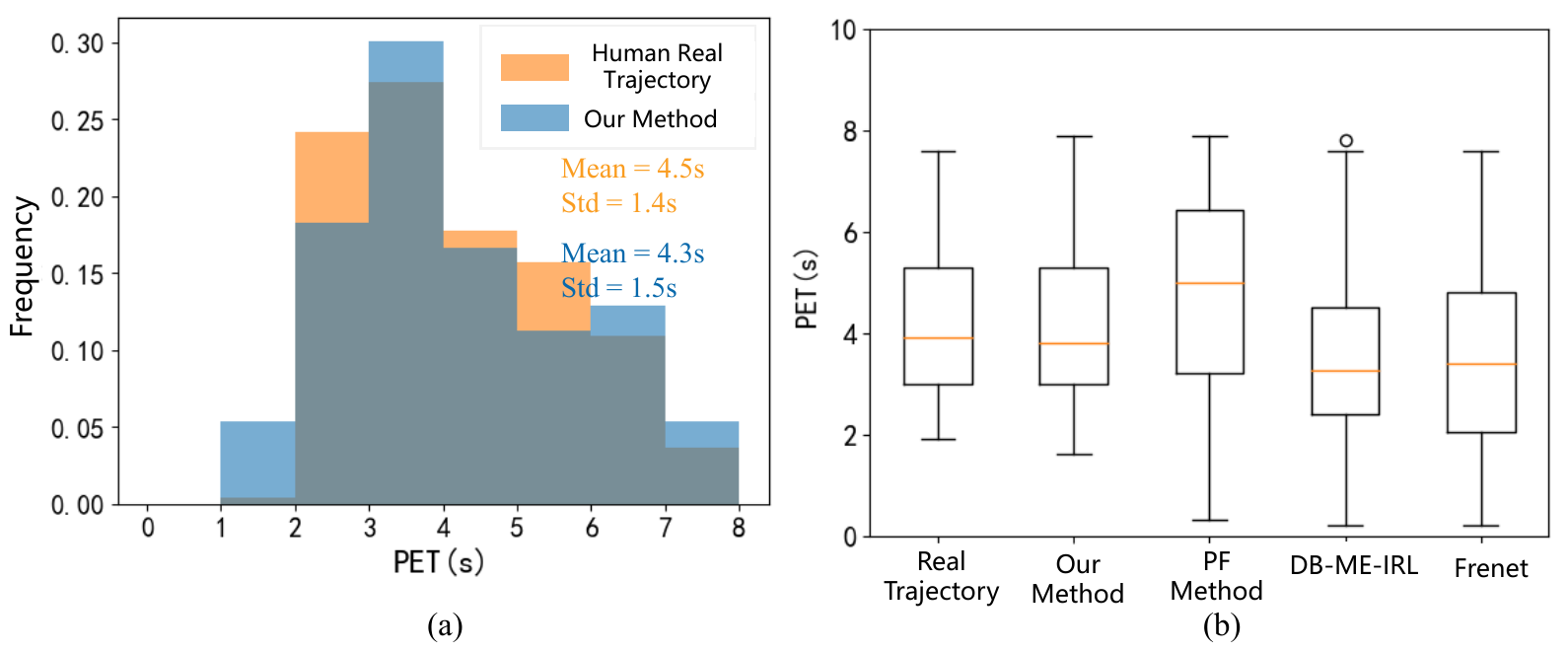}
    \caption{PET distribution of planning trajectory interaction process.}
    \label{fig:result_PET_distri}
\end{figure}

PET, a critical measure of conflict severity during left turns, was also evaluated. The average PET for our method was 4.3s, compared to the real trajectory average of 4.5s (Fig. \ref{fig:result_PET_distri}). Our method's PET performance was the most favorable, with the DB-ME-IRL and Frenet methods showing a marked decrease in PET, implying a higher risk in trajectory interaction safety.

\subsubsection{Assessment of Computational Efficiency and Learning Efficacy}
The study also aimed to evaluate the enhanced computational efficiency of our trajectory planning algorithm, particularly focusing on how the decision expectation space constraint influences this efficiency. Key parameters for this assessment included the average number of candidate trajectories, average computation time (incorporating both feature computation and trajectory search), and the overall learning performance. Comparative experiments were conducted against the DB-ME-IRL method, and the outcomes are tabulated in Tab. \ref{table:calculate_efficiency_compare}. The experiments utilized a computing setup with an i7-8700 processor and 16GB of memory. Our method showed a significant reduction in both the average number of candidate trajectories (by 52.5\%) and computation time (by 41.1\%) after applying the expected trajectory space constraint.

\begin{figure}
    \centering
    \includegraphics[width=0.8\linewidth]{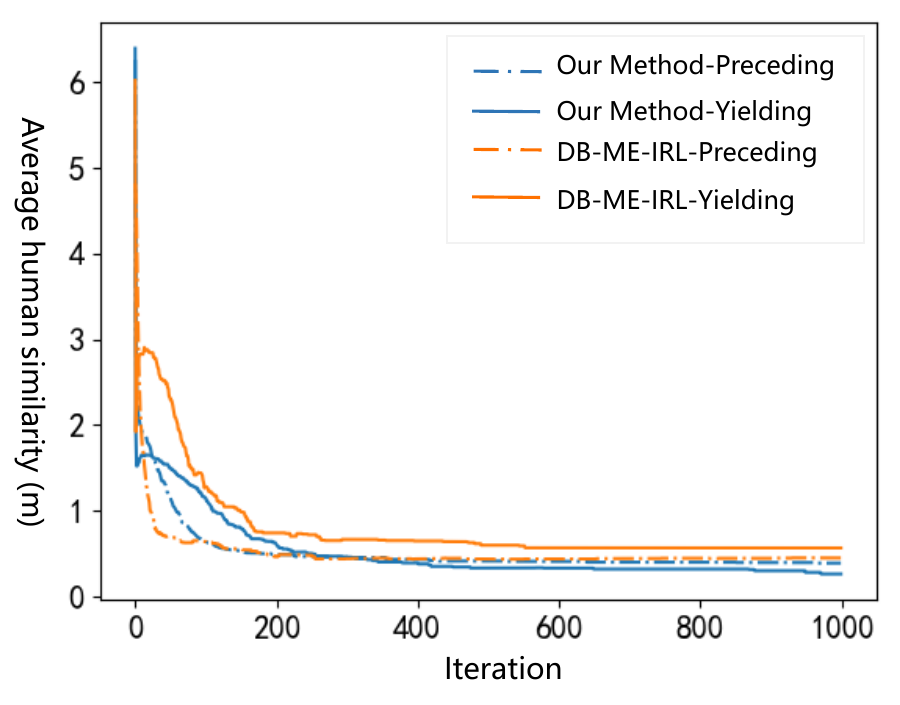}
    \caption{Comparison of training process effects between DB-ME-IRL and our method.}
    \label{fig:IRL_effect_compare}
\end{figure}

Furthermore, the trajectory learning performance was assessed using the Average Human Likeness (AHL) index. Both our method and the DB-ME-IRL method underwent training over 1000 iterations. The resultant AHL values for our method in priority and yielding decision scenarios were 0.39 and 0.26, respectively, demonstrating a marked improvement of 13\% and 54\% in learning performance compared to the DB-ME-IRL method. These results are comprehensively presented in Tab. \ref{table:calculate_efficiency_compare} and Fig. \ref{fig:IRL_effect_compare}.

\begin{table*}
    \centering
    \caption{Trajectory Planning Efficiency and Effect Comparison.}
    \label{table:calculate_efficiency_compare}
    \begin{tabular}{>{\centering\arraybackslash}p{3cm} >{\centering\arraybackslash}p{3cm} >{\centering\arraybackslash}p{3cm} >{\centering\arraybackslash}p{3cm} >{\centering\arraybackslash}p{3cm}}
    \toprule
        Method & Average Number of Candidate Trajectories & Average Calculation Time & Learning Effect (Preceding) & Learning Effect (Yielding)\\ 
    \midrule
        Our Method & 273 & 0.079s & 0.39 & 0.26 \\ \hline
        Decision-based ME-IRL & 581 & 0.134s & 0.45 & 0.57 \\
    \bottomrule
    \end{tabular}
\end{table*}

\section{Evaluation on the Human-in-loop Driving Platform}
Although low-cost, efficient, and rapid experiments can be conducted in simulated environments, vehicles in simulators typically exhibit fixed driving styles and behavior patterns, lacking the ability to perceive and understand socio-interactional cues. To validate the algorithm's socio-interactional capabilities, in this section, we establish a Human-in-Loop driving simulation platform and design simulation experiments involving human drivers to assess the algorithm's interpretability of intent and timing of intent clarification.

\subsection{Platform Construction and Experimental Design}
\begin{figure*}
    \centering
    \includegraphics[width=0.6\linewidth]{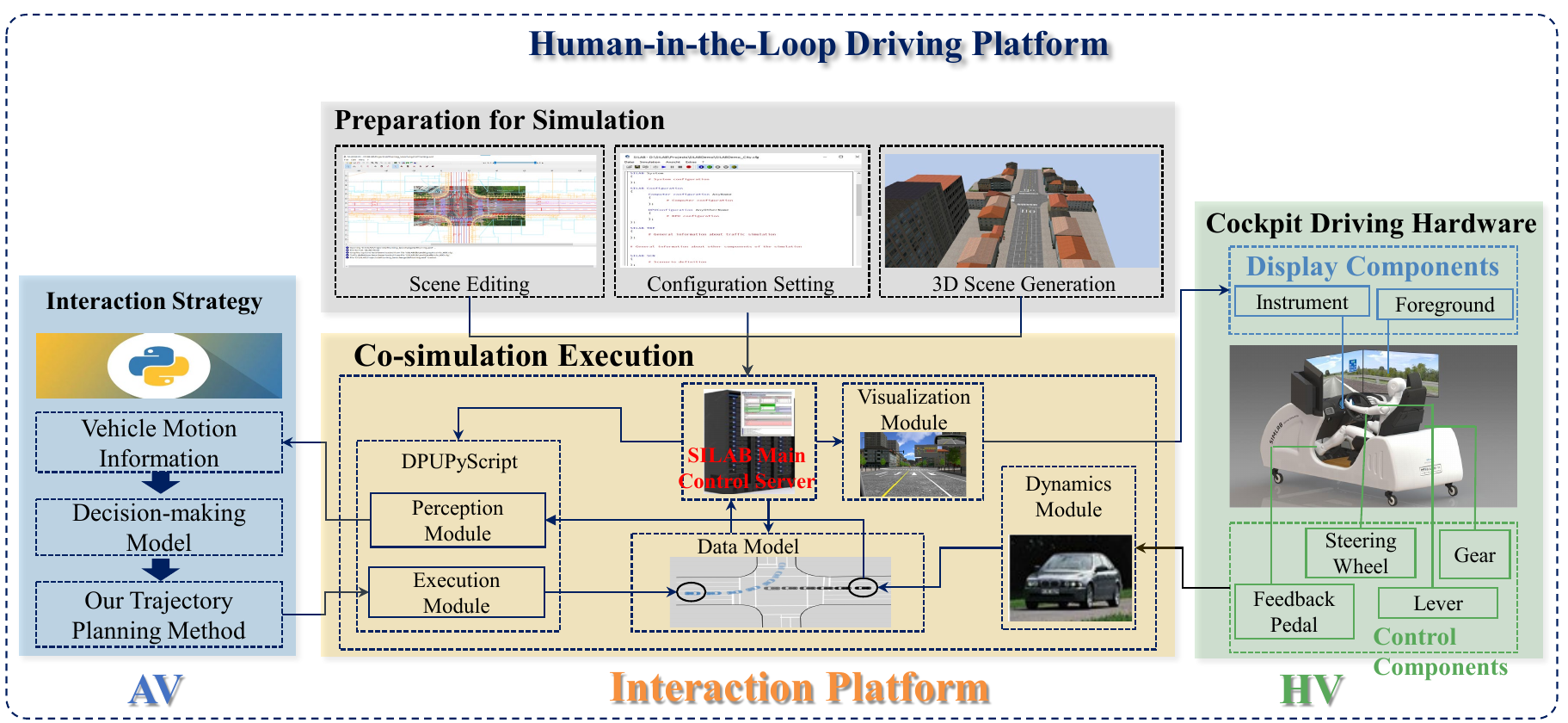}
    \caption{The human-in-loop driving platform in our experiment.}
    \label{fig:human-in-loop driving platform}
\end{figure*}

The Human-in-Loop driving simulation platform was established for our study, as showcased in Fig.~\ref{fig:human-in-loop driving platform}. This platform comprises three integral components: the hardware setup, software framework, and a series of interaction strategies. We recruited fourteen participants, all holding valid driver's licenses with a minimum of one year's driving experience, to partake in our experiments. The simulated environment replicated an intersection with east-west directions featuring bidirectional four-lane roads and north-south directions having bidirectional two-lane roads. The experimental setup involved left-turning vehicles (operated by our algorithm) entering from the west and straight-going vehicles (managed by participants) from the east.

The ME-IRL algorithm was utilized as the baseline for comparison, with high-level decision-making in both our and the baseline algorithm derived from a decision tree algorithm~\cite{ma2017two}. Each participant underwent 20 different scenarios, resulting in 280 sets of interaction data.

A series of subjective perception surveys were conducted before, during, and after the experiments to assess participants' trust in current AVs, feedback on AV interaction strategies, and post-experiment reflections on strategy acceptability and trust alterations. These surveys, detailed in Tab.~\ref{tab:subjective_evaluation}, employed a 5-point Likert scale, apart from decision consistency measured in binary terms during the experiments. Further details on the simulation platform and experiment design are available at the specified link.\footnote{See \url{https://drive.google.com/drive/folders/1hMP1QvL8jq-jti_rTme5aG1GQ7POm6K1?usp=drive_link}}

\begin{table*}
    \centering
    \caption{Subjective Evaluation using a 5-Point Likert Scale at Different Experimental Stages.}
    \label{tab:subjective_evaluation}
    \begin{tabularx}{1.0\textwidth}{lXl}
        \toprule
        \textbf{Experimental Phase} & \textbf{Subjective Perception Question} & \textbf{5-Point Likert Scale} \\
        \midrule
        Before Experiment & \textbf{Question about Trust Level} & Not at all Trusting — Completely Trusting \\
        & During the current stage of driving and interaction with the AV, do you trust the decision-making behavior of the AV? & \\
        \midrule
        During Experiment & \textbf{Question about Interpretability} & Not at all Clear — Completely Clear \\
        & During this driving process, do you think the left-turning vehicle (AV) expresses its intent clearly? & \\
        & \textbf{Question about Timing of Intent Clarification} & Very Late — Very Early \\
        & During the interaction process, at which stage can you fully understand the intent of the left-turning vehicle? & \\
        \midrule
        After Experiment & \textbf{Question about Approval Level} & Not Necessary — Very Necessary \\
        & Do you think it is necessary for the AV to showcase its intent through different actions? & \\
        & \textbf{Question about Trust Level} & Not at all Helpful — Extremely Helpful \\
        & Does this socio-interaction strategy help improve your trust in the AV? & \\
        \bottomrule
    \end{tabularx}
\end{table*}

\subsection{Experiment Result}
The main focus of our analysis was on the social performance of the algorithm, particularly in terms of interpretability of intent and the timing of intent clarification.

\subsubsection{Interpretability of Intent}
\begin{figure}
    \centering
    \includegraphics[width=1\linewidth]{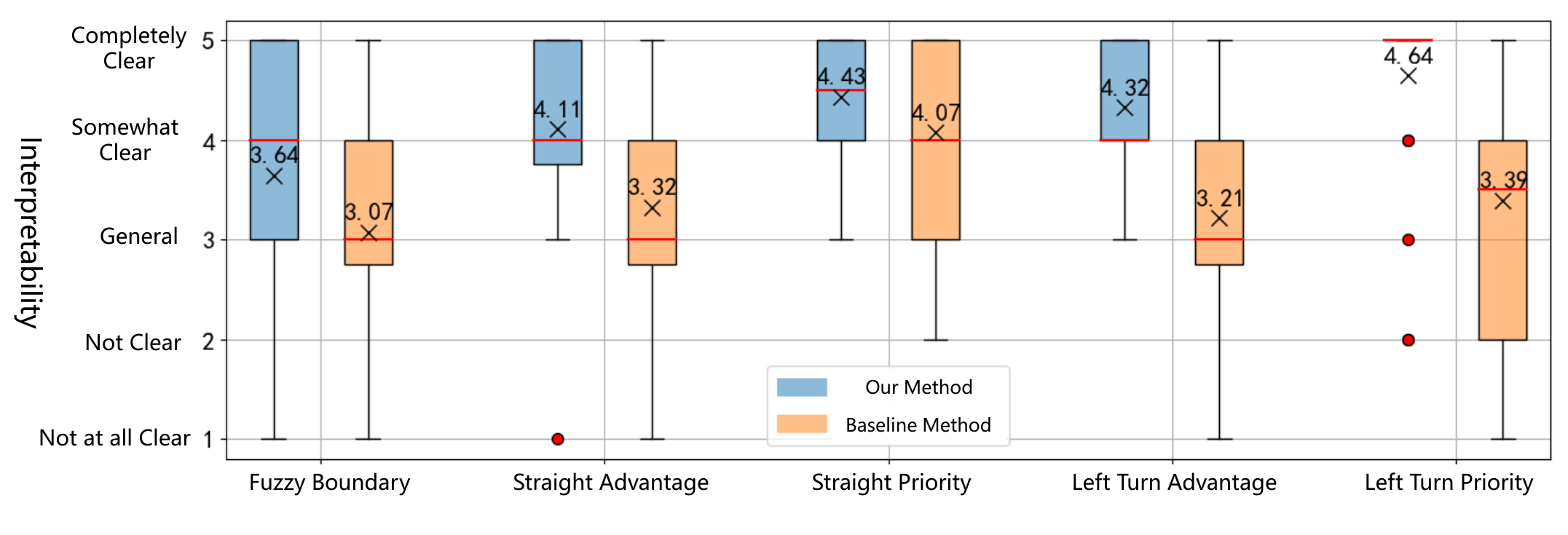}
    \caption{Subjective evaluation of intent interpretability for different interaction strategies in various precedence advantage scenarios.}
    \label{fig:human interpretability evaluation}
\end{figure}

The interpretability of our strategy and comparative strategies was subjected to statistical analysis across various priority scenarios, as depicted in Fig.~\ref{fig:human interpretability evaluation}. The Wilcoxon non-parametric tests were used to compare subjective perception scores for interpretability, followed by Pearson correlation coefficient analysis to assess the impact of different scenarios on participants' perceptions. The results are summarized in Tab.\ref{tab:interpretability_results}. In the table,the values outside and inside the brackets in the Median and Mean columns represent the respective values for our strategy and the comparative strategy.

Our strategy consistently achieved higher interpretability scores than comparative strategies across all scenarios, as shown in Fig.\ref{tab:interpretability_results}, significant differences were noted in interpretability between our strategy and the comparative one, particularly in scenarios involving straight-going and left-turn priorities. Our strategy was found to have a more substantial impact on interpretability, especially in left-turn priority and advantage scenarios. The overall analysis indicated a median interpretability score of 5.0 ("very clear") and a mean score indicating a clarity level between "somewhat clear" and "very clear."

\begin{table*}
    \centering
    \caption{Descriptive Statistical Results of Subjective Evaluation on the Interpretability of Intent for Different Strategies. }
    \label{tab:interpretability_results}
    \begin{tabular}{lcccccc}
        \toprule
        \textbf{Scenario} & \textbf{Median (Mdn)} & \textbf{Mean} & \textbf{Test Statistic (W)} & \textbf{p-value} & \textbf{Effect Size (r)} \\
        \midrule
        Fuzzy Boundary (n=28) & 4.0 (3.0) & 3.64 & 85.0 & 0.057 & 0.24 \\
        Straight Advantage (n=28) & 4.0 (3.0) & 4.11 & 37.5 & 0.010 & 0.35 \\
        Straight Priority (n=28) & 4.5 (4.0) & 4.43 & 24.0 & 0.058 & 0.22 \\
        Left Turn Advantage (n=28) & 4.0 (3.0) & 4.32 & 22.5 & 0.001 & 0.51 \\
        Left Turn Priority (n=28) & 5.0 (3.5) & 4.64 & 22.5 & 0.000 & 0.52 \\
        All Scenarios (n=140) & 5.0 (3.0) & 4.23 & 847.5 & 0.000 & 0.36 \\
        \bottomrule
    \end{tabular}
    \smallskip
\end{table*}

\subsubsection{Timing of Intent Clarification}
Alongside interpretability, the timing of intent clarification was also a key focus. Earlier clarification of intent by participants suggested a more explicit and timely expression of social intent by our interaction strategy.

\begin{figure}
    \centering
    \includegraphics[width=1\linewidth]{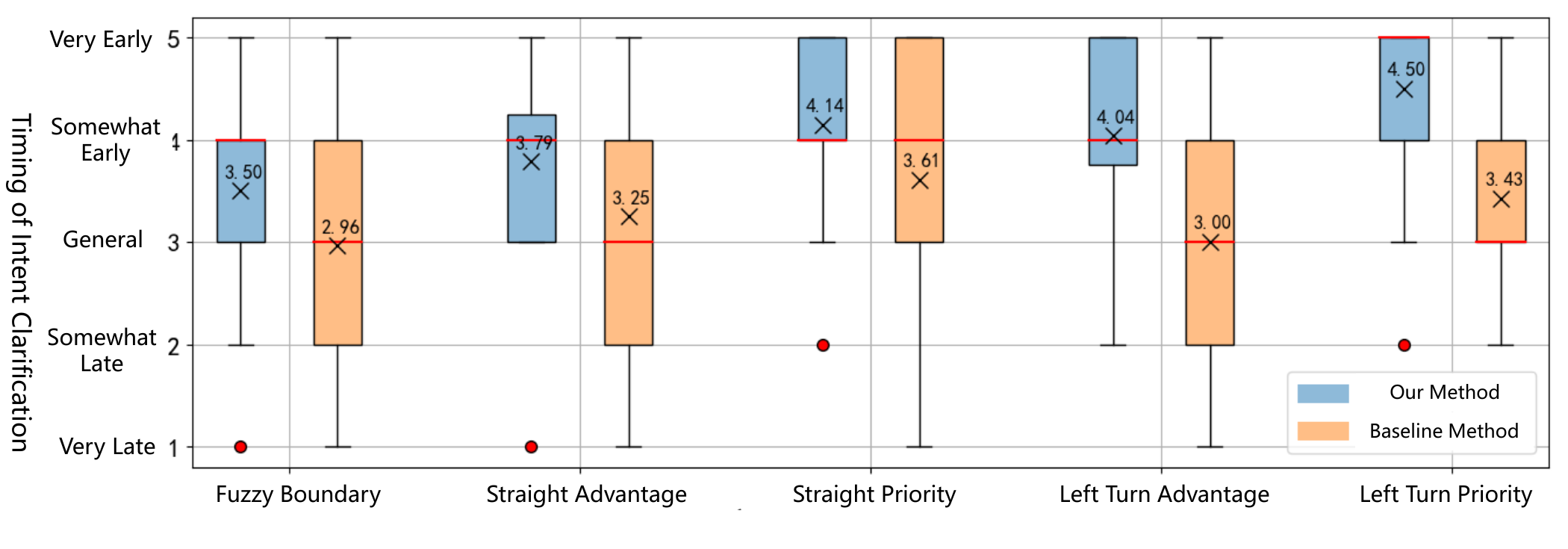}
    \caption{Subjective evaluation of timing clarity of intent for different interaction strategies in various precedence advantage scenarios.}
    \label{fig:human_time_clarification_evaluation}
\end{figure}

Statistical comparisons of our strategy and the comparative strategy regarding intent clarification timing were conducted across different scenarios, as illustrated in Fig.\ref{tab:timing_results} and Fig.~\ref{fig:human_time_clarification_evaluation} indicates that our strategy generally allowed for earlier intent clarification across all scenarios. Wilcoxon tests showed significant differences in all scenarios, confirming that our strategy facilitated earlier and more explicit expression of intent. Pearson correlation analysis further suggested that our strategy most significantly impacted scenarios with left-turn advantage, followed by left-turn priority. In sum, our strategy achieved a median score of 4.0 ("somewhat early") in intent clarification timing, with an average indicating a range between "somewhat early" and "very early," marking a noteworthy improvement over the comparative strategy.


\begin{table*}
    \centering
    \caption{Descriptive Statistical Results of Timing of Intent Clarification for Different Strategies. }
    \label{tab:timing_results}
    \begin{tabular}{lcccccc}
        \toprule
        \textbf{Scenario} & \textbf{Median (Mdn)} & \textbf{Mean} & \textbf{Test Statistic (W)} & \textbf{p-value} & \textbf{Effect Size (r)} \\
        \midrule
        Fuzzy Boundary (n=28) & 4.0 (3.0) & 3.50 & 59.0 & 0.044 & 0.23 \\
        Straight Advantage (n=28) & 4.0 (3.0) & 3.79 & 52.5 & 0.022 & 0.25 \\
        Straight Priority (n=28) & 4.0 (4.0) & 4.14 & 21.0 & 0.009 & 0.25 \\
        Left Turn Advantage (n=28) & 4.0 (3.0) & 4.04 & 6.0 & 0.000 & 0.48 \\
        Left Turn Priority (n=28) & 5.0 (3.0) & 4.50 & 19.5 & 0.000 & 0.52 \\
        All Scenarios (n=140) & 4.0 (3.0) & 4.00 & 736.0 & 0.000 & 0.33 \\
        \bottomrule
    \end{tabular}
    \smallskip
\end{table*}

\section{Conclusion}
\label{sec:5}
In an endeavor to narrow the divide between AVs and HVs and ensure that AVs can implicitly convey social intent in a manner understandable to HVs in mixed-traffic scenarios, we have introduced an innovative framework for socially-compliant trajectory planning that is robust in implicit intent expression at the unprotected left-turn scenarios.

Our proposed framework is organized into three components: trajectory generation, trajectory evaluation, and trajectory selection. The experimental results substantiate the efficacy of our framework, demonstrating a strong resemblance to actual human trajectories, considerable enhancements in intent expression, safety, and efficiency, along with improved computational efficiency and learning outcomes. Our method shows a 77\% match with the actual trajectory distribution, an average offset of 85\% from the real trajectory, an average travel time of 7.4 seconds within the intersection, and a decrease in the average computation time by 41.1\%.
Concurrently, experiments conducted within the simulation platform and human-in-loop driving platform unequivocally showcase the effectiveness and precedence of the algorithm.

For future research, our research focus will expand the applicability of our trajectory planning methodology to include a broader range of interactive scenarios. Furthermore, we aim to confirm the effectiveness and scalability of our methodology through driving simulation experiments and real-vehicle interactions.


\bibliographystyle{IEEEtran} 
\bibliography{reference}

\vfill

\end{document}